# Combined Person Classification with Airborne Optical Sectioning

Indrajit Kurmi, David C. Schedl, and Oliver Bimber

***Abstract*—Fully autonomous drones have been demonstrated to find lost or injured persons under strongly occluding forest canopy. Airborne Optical Sectioning (AOS), a novel synthetic aperture imaging technique, together with deep-learning-based classification enables high detection rates under realistic search-and-rescue conditions. We demonstrate that false detections can be significantly suppressed and true detections boosted by combining classifications from multiple AOS – rather than single – integral images. This improves classification rates especially in the presence of occlusion. To make this possible, we modified the AOS imaging process to support large overlaps between subsequent integrals, enabling real-time and on-board scanning and processing of groundspeeds up to 10 m/s.**

## I. Introduction

SYNTHETIC aperture (SA) sensing is a widely used technique for emulating wide aperture sensors where they would be physically infeasible. It acquires individual signals of multiple or a single moving small-aperture sensor to computationally combine them to improve, for instance, resolution, depth of field, frame rate, contrast, or signal-to-noise ratio. SA sensing has been utilized with diverse sensors in a wide range of applications, such as radar [1][2][3] (obtaining weather-independent images and reconstructing geospatial depth), radio telescopes [4][5] (observing large celestial phenomena in outer space), microscopes [6] (reconstructing a defocus-free 3D volume using interferometry), sonar [7][8][9][10] (generating high-resolution mappings of underwater objects and seafloors), ultrasound [11][12] (non-intrusive intravascular 3D imaging), laser [13][14] (earth observation utilizing shorter wavelengths using LiDAR), and optical imaging [15][16][17][18][19][20][21][22] (acquiring structured light fields with large camera arrays for various post-processing steps, such as refocusing, computation of virtual views with maximal synthetic apertures, and varying depth of field).

Airborne Optical Sectioning (AOS) [23][24][25][26][27][28][29][30][31] is an effective wide-synthetic-aperture aerial imaging technique that can be deployed using camera drones. It allows virtual mimicking of a wide aperture optic of the shape and size of the scan area (possibly hundreds to thousands of square meters) that generates images of an extremely shallow depth of field above an occluding structure, such as a forest. These images are computed by integrating regular single pictures that are captured by the drone, and allow optical slicing through dense occlusion (caused by leaves, branches and bushes). In each slice, AOS can reveal targets, such as artifacts, objects, wildlife, or persons, which would remain occluded for regular cameras.

Compared to alternative airborne scanning technologies, such as LiDAR [32][33][34] and synthetic aperture radar [1][2][3], AOS is cheaper, wavelength independent, and offers real-time computational performance for occlusion removal. We have applied AOS within the visible [23] and the thermal [26] spectra, and demonstrated its usefulness in archeology [23], wildlife observation [27], and search and rescue (SAR) [30][31]. By employing the randomly distributed statistical model in [25], we explained AOS' efficiency with respect to occlusion density, occluder sizes, number of integrated samples, and size of the synthetic aperture.

Recently, we have proven that classification of partially occluded persons in forests using aerial thermal images is significantly more effective when AOS is used to integrate single images before classification than when classification results of single images are combined [30]. We have also demonstrated the real-time application of AOS in fully autonomous and classification-driven adaptive SAR operations [31]. Currently, we achieve average precision scores of *86.0%-92.2%* under realistic conditions [30][31].

In this article we show how the chance of detecting persons can be increased by combining multiple classification results from strongly overlapping AOS integral images. Combined classification has been widely studied for improving overall classifier performance (improving accuracy, dealing with diverse and noisy datasets, etc.) [35][36][37][38][39][40][41]. Combination mechanisms can be used at various levels of the classification [42][43][44][45][46][47][48][49][50][51][52] (e.g., at the data, feature or decision levels). Decision-level combination is particularly popular due to its simplicity (no extensive knowledge of classifiers required) leading to lower complexity of the combination method. Most decision-level combination techniques are broadly categorized based on the output of the classifier (e.g., abstract, rank, and measurement). The most informative are measurement-based (also known as score-based) combination methods [36]. Approaches for measurement-based combination can be further categorized into adaptive [47][48][49][50] and non-adaptive methods [39][51[52]. Adaptive techniques are based mainly on evolution, artificial intelligence algorithms or fuzzy set theories, while non-adaptive combination techniques apply simple rules for combination, such as sum, product, maximum, and median [39][51][52]. No theoretical or empirical evidence of general superiority of any particular combination scheme exists, and even simple combination schemes have been shown to improve accuracy in various systems [40][41].

To guarantee real-time rates on low-performance on-device mobile processors, we consider non-adaptive, measurement-based combination techniques. We demonstrate in Sections IV and V that (compared to independent classifications in single integral images) the product of median and maximum confidence scores of combined classifications significantly suppresses false while boosting true detections. However, to enable a combined classification initially, the AOS imaging process must support large overlaps between subsequent integral images, which was not hitherto feasible. Sections III and VI discuss optimal AOS sampling parameters and how we achieve these overlaps. We start with a brief review of the AOS principle in Section II.

## II. Airborne Optical Sectioning

As shown in Fig. 1, AOS captures thermal radiation with a drone that samples forest within the range of a synthetic aperture (SA) at flying altitude. This results in multiple geotagged aerial thermal images and consequently in unstructured thermal light-field rays formed by image pixels and camera poses on the SA [53][54]. With known camera intrinsics, drone poses, and a representation of the terrain (e.g., a digital elevation model or a focal plane approximation [28]), each ray's origin on the ground can be reconstructed. Averaging all rays with the same origin results in a focused and widely occlusion-free integral image of thermal sources (e.g., persons) on the ground.

There is a statistical chance that a point on the forest ground is unoccluded by vegetation from multiple perspectives, as explained by the probability model in [25]. Thus, depending on density, more or fewer rays of a surface point contain information of random occluders, while others carry the constant signal of the target. Integrating multiple rays deemphasizes the occlusion signal and emphasizes the target signal. Since the remaining occlusion only lowers the contrast of the target [25], reliable classification of the target is possible – even under strong occlusion conditions [30].

AOS relies on the camera’s pose information while capturing images within a certain SA. In [30], the SA was a 2D sampling area, and precise pose estimation was achieved with computer-vision-based reconstruction techniques. This resulted in high-quality integral images, but was – for two reasons – not usable in time-critical applications, such as search and rescue: Sampling a 2D area led to long flying times, and precise computer-vision-based pose estimation was not possible in real time. All computations were carried out offline and after recording at predefined SA waypoints. This was significantly improved in [31] by sampling along short 1D SAs (linear flight paths) and by using instant but imprecise sensor readings from the drone (barometer altitude, non-differential GPS location, and compass orientation) for real-time computations directly on the drone. Despite the imprecision caused by lower-dimensional sampling and imprecise pose estimation, person classification performance was similar to that of 2D SA sampling with precise pose estimation [31].

To date, [30][31] person classification with AOS has only been achieved in discrete (non-overlapping) integral images, for which 30 single images were required at 1 m intervals before they could be integrated and classified after a 30 m flight distance and 30 s of flight time. Due to slow recording speed and long image transfer times from the camera to the drone’s processor, the ground surface covered did not overlap in multiple integral images. Consequently, the probability of detecting a person on the ground surface relied solely on a single classification chance. If, under unfavorable occlusion

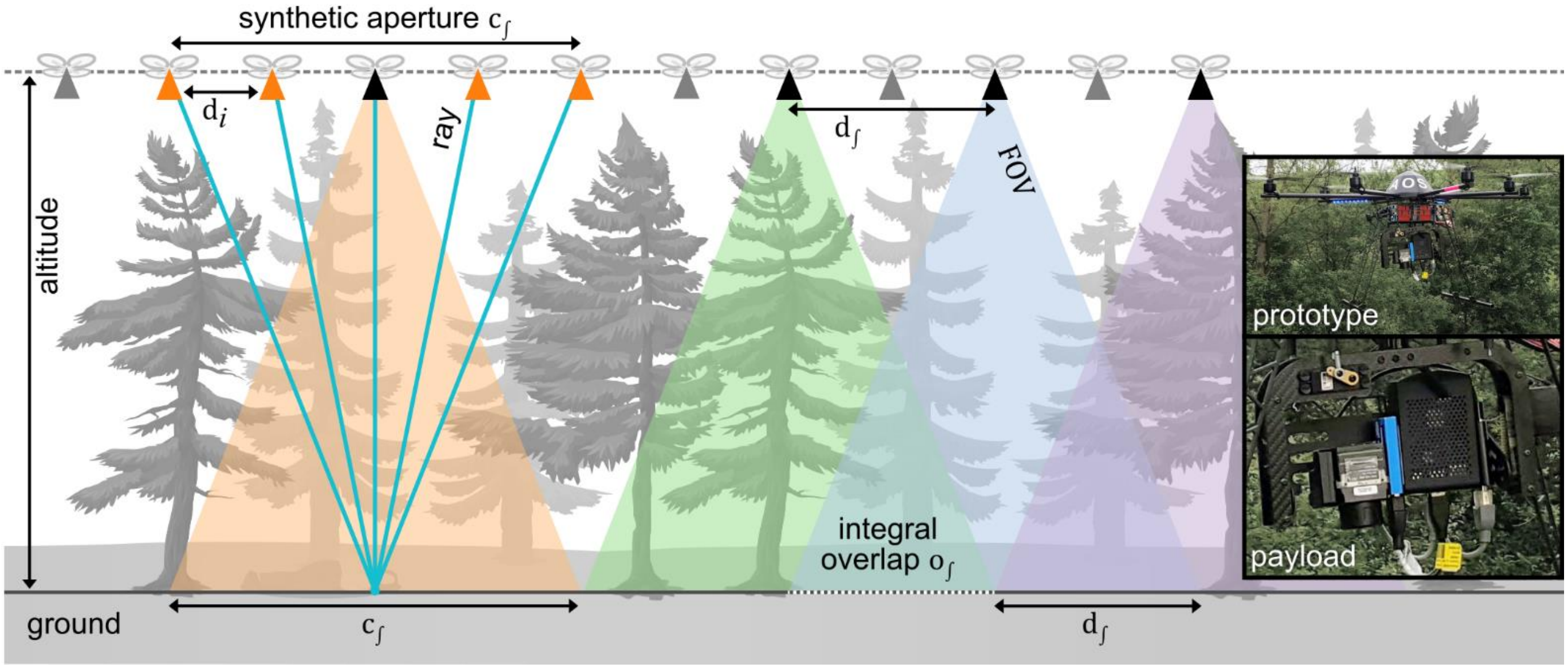

Fig. 1. Synthetic aperture sampling with Airborne Optical Sectioning: Sampling parameters and current drone prototype with payload (right).

conditions, a person was not detected in the corresponding integral image, they were never found.

In the sections below, we present the theoretical and practical foundations of continuous 1D synthetic aperture sampling (i.e., the real-time capturing, processing, and evaluation of integral images with a large ground surface overlap); it increases the chance of detecting a person proportionally to the amount of integral image overlap by combining multiple classification results of the same spot on the ground.

## III. Continuous 1D Synthetic Aperture Sampling

For continuous 1D synthetic aperture sampling, various sampling parameters, such as flying speed and altitude, imaging and processing speed, and camera field of view can be considered (cf. Figs. 1 and 2):

The distance between two subsequent integral images depends on the drone's flying speed $v_f$ [m/s] and the required processing time $t_p$ [s]:

$$d_\int = v_f \cdot t_p, \tag{1}$$

where $d_\int$ increases with flying speed and processing time. For instance, for a computational performance (including integral image computation and classification) of $t_p = 0.5$ s (i.e., two processing passes per second), $d_\int$ is 0.5 m, 2 m, 3 m, and 5 m for $v_f$ of 1 m/s, 4 m/s, 6 m/s, and 10 m/s, respectively.

The ground coverage of each integral image depends on the flight altitude $h$ [m] and the camera's field of view $FOV$ [°]:

$$c_\int = 2 \cdot h \cdot \tan(FOV/2). \tag{2}$$

Thus, for $h = 35$ m above ground level (AGL) and a $FOV$ of 43°, for example, $c_\int$ is 27.6 m (in one direction).

The amount of integral image overlap (i.e., how often the same surface point is covered by subsequent integral images) is:

$$o_\int = c_\int / d_\int. \tag{3}$$

At a $FOV$ of 43°and an increasing $d_\int$ of 0.5 m, 2 m, 3 m, and 5 m, for instance, $o_\int$ decreases: 55.2, 13.8, 9.2, and 5.5. In principle, $o_\int$ represents the number of times the same person can be detected while being scanned. Note that $o_\int$ should not fall below 1, as this results in imaging gaps (ground portions not being covered at all).

The sampling density of integral images is the number of single images $N$ being integrated:

$$D_\int = N = c_\int / d_i, \tag{4}$$

where $d_i$[m] is the sampling distance of single images. It correlates with the efficiency of occlusion removal, and has an upper limit, as discussed in [25]. Note that $t_p$ also increases proportionally with $N$. Furthermore, the rate at which single images are recorded should be equal to or higher than the rate at which integrals are computed ($d_\int \geq d_i$). If $d_\int < d_i$, subsequent integral images will not change.

The integration time (i.e., the time required to capture $N$ images that are combined into one integral image) is:

$$t_\int = c_\int / v_f = d_i \cdot D_\int / v_f. \tag{5}$$

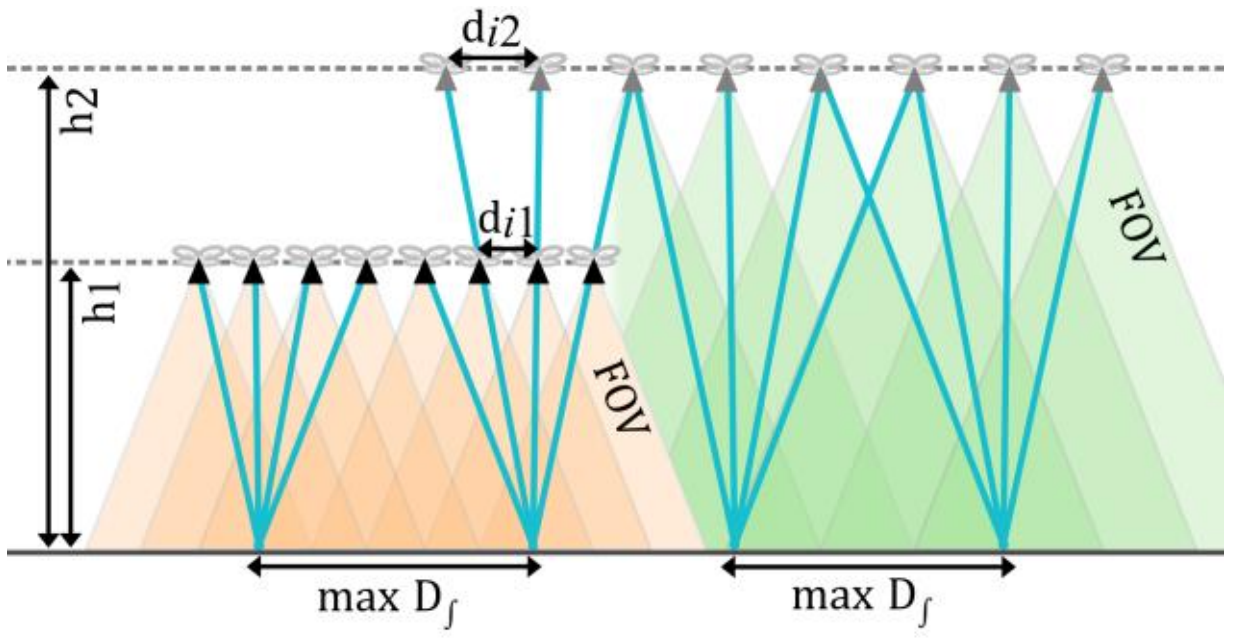

Fig. 2. Sampling distances $d_i$ of single images with the same image disparity and $FOV$ at two different flight altitudes $h_1$ and $h_2$. Note that a higher altitude does not result in more coverage on the ground, with a maximum $D_\int$.

For $c_\int$=27.6 m, $t_\int$ is 27.6 s, 6.9 s, 4.63 s, and 2.8 s for flying speeds $v_f$ of 1 m/s, 4 m/s, 6 m/s, and 10 m/s. Note that a large $t_\int$ is unfavorable for the detection of fast moving persons, as they introduce motion blur to the integral images that might not be classified correctly.

The sampling distances of single images that cause the same image disparity required for effective occlusion removal (as explained in [25]) at different flight altitudes (but at the same $FOV$) are linearly related (cf. Fig. 2):

$$d_{i1}/d_{i2} = h_1/h_2. \tag{6}$$

Thus, to match the occlusion removal efficiency of $d_{i1}$=1 m sampling distance at $h_1$=35 m, AGL requires a sampling distance of $d_{i2}$=28.6 m at $h_2$=1000 m AGL. Both synthetic aperture size and ground region covered scale proportionally to $h_1/h_2$ to achieve the same $D_\int$ at these two altitudes. Covering the same ground region at the same $D_\int$, however, requires the same scanning time for the same $v_f$, as an identical distance must be flown. Scanning at lower altitudes ($h_1 < h_2$) benefits from a $h_2/h_1$ times higher spatial sampling resolution compared to scanning from higher altitudes ($h_2 > h_1$), and from $h_2/h_1$ times more intermediate classification results for the same scanning distance, as $h_2/h_1$ times more single images are captured.

As explained in [25], the orthographically projected occlusion density $\widetilde{D}$ in single images depends statistically on density and size of the occluders (i.e., density of forest and size of branches, trunks, leafs, etc.) and the height of the occlusion volume (i.e., the height of trees). Here, the assumption of an orthographic projection implies an orthographic viewing angle $\alpha = 0$ with respect to the synthetic aperture (SA) plane (i.e., looking straight down at the forest ground). More oblique viewing angles ($\alpha > 0$), however, result in an increase in projected occlusion density due to the longer imaging distance from the SA plane through the occlusion volume to the forest ground (see proof in appendix):

$$\widetilde{D_\alpha} = 1 - (1 - \widetilde{D})^{1/cos(\alpha)}. \tag{7}$$

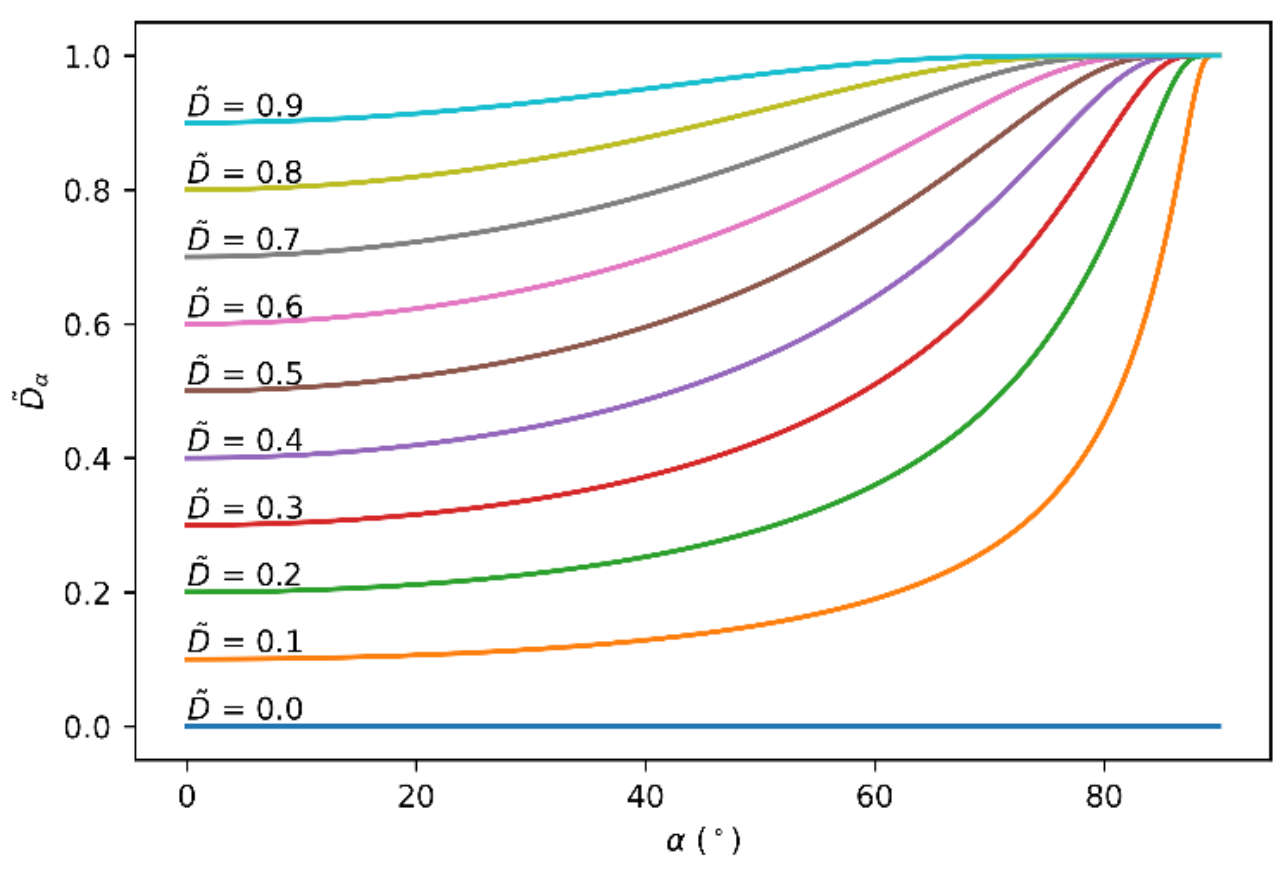

Fig. 3. Increase in projected occlusion density ($\widetilde{D_\alpha}$, $\tilde{D}$ for $\alpha = 0$) with changing viewing angle $\alpha$.

As illustrated in Fig. 3, large viewing angles (and consequently cameras with a large $FOV = 2 \cdot \alpha$) are inefficient for occlusion removal. In the case of forests, oblique viewing angles would additionally cause larger projected occluder sizes, since side views of tree trunks project to larger footprints than top-down views. Larger occluder sizes, however, would require even larger sampling distances for efficient occlusion removal [25].

Considering all the above sampling parameters, the following conclusions can be drawn:

(1) A larger $FOV$ is beneficial only up to a certain degree (when occlusion removal becomes seriously inefficient because viewing angles are too oblique).
(2) For a given flying speed, higher flight altitudes have no effect on scanning time (i.e., the time needed to cover a certain forest range), but lead to lower spatial sampling resolution on the ground. Flight altitudes should therefore be chosen to be as low as possible.
(3) A higher flying speed decreases not only scanning time, but also the integral image overlap (and thus the number of times a person can be detected correctly). A suitable trade-off between flying speed and detection probability must be chosen. This might depend on the amount of occlusion (i.e., slower flights for denser forests).
(4) Faster imaging speeds and shorter processing times are always beneficial, as both increase integral image overlap and allow faster flying speeds.

## IV. Combined Classification

Continuous computation of integral images and combined classifications within them increase the chance that a person is detected correctly by a factor of $o_\int$ compared to single classifications of non-overlapping integral images, as in [30][31].
Many real-time object-detection algorithms, such as YOLO [55], output classification results as axis-aligned bounding boxes (AABBs) together with detection confidence scores. For classification combination, we project all AABBs of all individual integral images to a common coordinate system that is defined by the DEM of the ground surface. Thus, for a single discrete point on the ground surface we collect a maximum of $o_\int$ overlapping AABBs and combine their individual confidence scores to a single score. This results in a DEM-aligned confidence map (initially filled with zero values). To distinguish between false and true detections on a per-pixel basis, each entry in this confidence map can be thresholded after $o_\int$ new image samples past its first appearance within the drone's FOV.
Our hypothesis is that true detections often (but not always) project higher scores to the same spatial location, while false detections project predominantly (but not exclusively) lower scores to more randomly distributed locations. Thus, combining the projected scores should emphasize true and suppress false detection scores, and consequently separate the score ranges of these two groups more clearly than in single integral image classifications. This will in turn allow better thresholding and thus improve overall classification performance.
Confidence scores can be combined by statistical or mathematical methods. We evaluated median, maximum, and the product of median and maximum, and present the results in Section V.

## V. Results

To evaluate our combination strategies, we performed 1D SA test flights at various constant speeds ($v_f$=4, 6, and 10 m/s) from 30 m AGL above unoccluded (open field) terrain, and 37 m above dense forests (conifer and broadleaf forest). Integral images were computed from $D_\int = N = 30$ single images recorded at a video rate of $t_i = 0.33$ s (30 fps) and a $FOV$ of 43.10°. The processing speed achieved was around $t_p = 0.5$ s.
Details on the hard- and software used, the new AOS image acquisition that achieves fast flying speeds and large integral image overlaps, the improved integral image computation that applies deferred rendering to reduce processing time, and the implementation of the person classification are provided in Section VI.
Fig. 4 illustrates the test sites and computed single integral images along the 1D SA flight paths together with the ground-truth labels of persons on the ground. As predicted by (3), the same person appears in $o_\int$ integral images for different $v_f$ (14 vs. 13.8 for 4 m/s, 9-10 vs. 9.2 for 6 m/s, and 5-6 vs. 5.2 for 10 m/s). Note that slight variations are caused by different transition angles through the $FOV$ and the fact that the $FOV$ differs slightly for horizontal/vertical and diagonal image axes.
Fig. 5 shows, for each flight, the probability maps of single integral images and combined classification results. For the open-field flights and single integral images, high confidence scores were obtained at the ground-truth positions, while low-score detections could easily be filtered out by a distinguished confidence threshold. This, however, was not the case for occlusion (especially during faster flights), where classification performance dropped significantly, and finding a proper confidence threshold to distinguish between false and true detections became increasingly difficult.
The results of maximum, median, and maximum·median combined classifications are shown in the last three columns of Fig. 5. While the maximum emphasized true but also aggregated false detections, and the median suppressed false but also reduced true detections, the maximum·median best suppressed false and emphasized true detections. That the score ranges between true and false detections can be separated best by a threshold with a maximum·median combination is also illustrated by the plots of confidence-score-ordered detections shown in Fig. 6. A steep gradient supports distinct confidence-score thresholds. Here, maximum·median outperformed a maximum and median combination as well as single integral image classification (without combination). Table 1 compares the maximum confidence scores of false detections with the minimum confidence scores of true detections. Their ratio represents the degree of separation between both groups (and how well confidence thresholds can be chosen). A value below 1 indicates false classifications, which was always the case for our forest flights and classification in single integral images. All combination methods increased separation (ratio > 1), and the maximum·median method performed best.

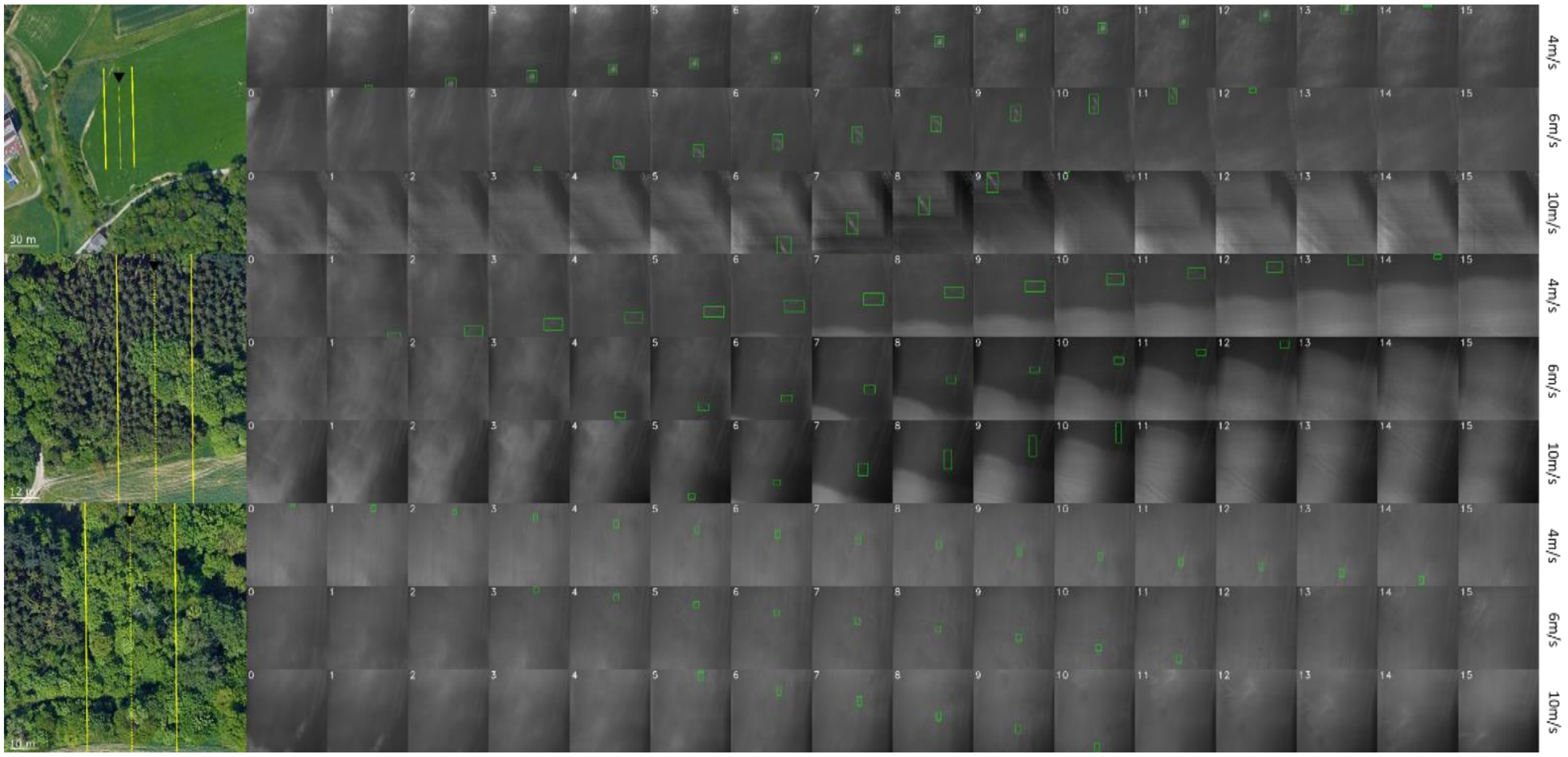

Fig. 4. Test sites, flight directions and coverages (left) at constant flying speeds of $v_f = 4$, 6, and 10 m/s over unoccluded (open field) terrain (top rows), over conifer forest (middle rows), and over broadleaf forest (bottom rows). Resulting integral images (right) reveal the appearance of the same person (bounding boxes indicate the manually labeled ground-truth appearances) of $o_f = 13.8/14$ for 4 m/s, 9.2/9-10 for 6 m/s, and 5-6/5.5 for 10 m/s. Note that the shape of the ground-truth bounding boxes varies slightly due to mis-registrations in the integral images. The GPS coordinates of the test sites are: 48°20'08.4"N, 14°19'34.6"E (open field), 48°19'58.1"N 14°19'48.1"E (conifer forest), 48°19'59.8"N 14°19'52.2"E (broadleaf forest).

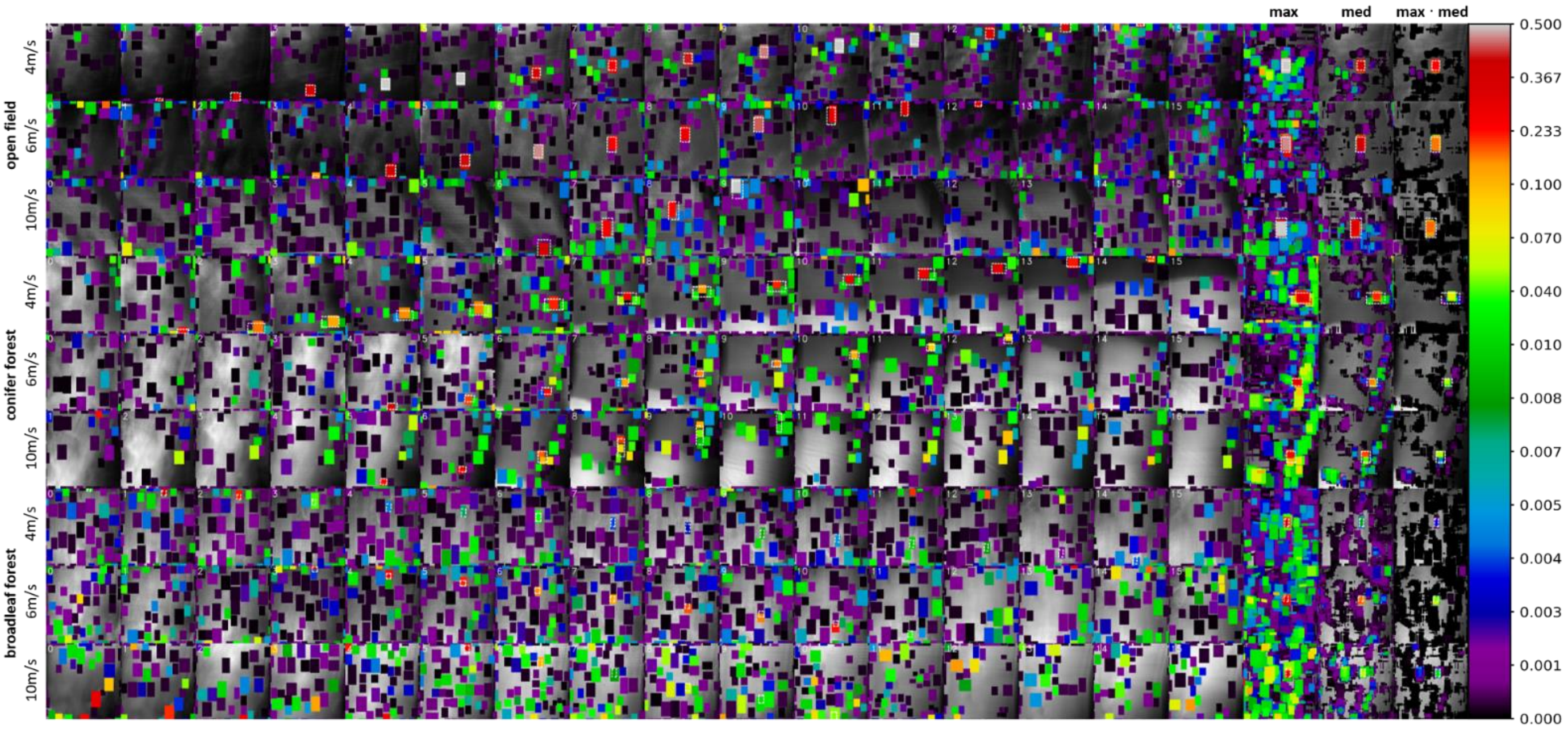

Fig. 5. Probability maps of single integral images (first 16 columns) and combined classification results (last three columns: maximum, median and maximum·median) for all test flights. Detections are indicated with AABBs, and confidence scores are color coded (see color bar on the right).

## VI. Methods

We utilized an octocopter (MikroKopter OktoXL 6S12, two LiPo 4500 mAh batteries, 4.5 kg;) that carried the following payload (cf. Fig. 1): a FLIR Vue Pro thermal camera (9 mm fixed focal length lens, 7.5 μm to 13.5 μm spectral band, 14-bit non-radiometric, 118 g), a Flir HDMI and power module providing HDMI video output from the camera (640x480 @30 Hz; 15 g), a video capture card (Basetech, 640x480 @30 Hz, 22 g), a single-board system-on-chip computer, SoCC (RaspberryPi 4B, 8 GB RAM, 65 g), an LTE communication hat (Sixfab 3G/4G & LTE base hat and a SIM card, 35 g), and a Vision Processing Unit, VPU (Intel Neural Compute Stick 2, 30 g). The payload (350 g in total) was mounted on a rotatable gimbal, and was positioned to keep the camera pointing downwards during flight.

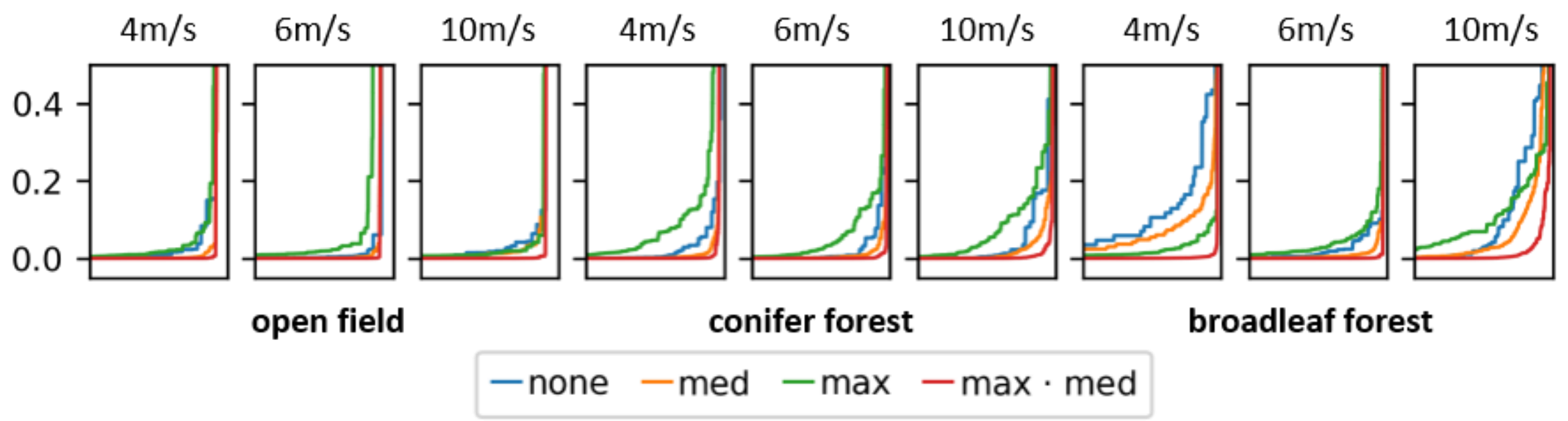

Fig. 6. Plotted confidence scores (y-axis) of all detections (x-axis) in low-to-high order (from left to right). The steepest gradient, which supports the best defined confidence-score thresholding, was always achieved with maximum·median combination. While the difference to other combination methods (and to no combination (none), that is, single integral image classification) was low without occlusion (open field), it was significant in the presence of occlusion (conifer and broadleaf forest).

Our software was implemented in Python and runs on the SoCC, where various sub-processes, such as drone communication, imaging acquisition, and image processing (including integral image computation and classification), run in parallel to make use of the SoCC's multi-core capabilities.

TABLE I
CONFIDENCE SCORES

| | | single | max | median | max · median |
|---|---|---|---|---|---|
| open field | 4m/s | 0.274/0.119<br>2.302 | 0.617/0.119<br>5.184 | 0.380/0.015<br>25.333 | 0.235/0.001<br>235.0 |
| | 6m/s | 0.249/0.099<br>2.515 | 0.471/0.099<br>4.757 | 0.365/0.015<br>24.333 | 0.172/0.0002<br>860.0 |
| | 10m/s | 0.241/0.033<br>7.303 | 0.565/0.033<br>17.121 | 0.313/0.033<br>9.484 | 0.177/0.001<br>177.0 |
| conifer forest | 4m/s | 0.122/0.147<br>0.829 | 0.305/0.147<br>2.074 | 0.221/0.092<br>2.402 | 0.067/0.008<br>8.375 |
| | 6m/s | 0.084/0.157<br>0.535 | 0.235/0.157<br>1.496 | 0.144/0.027<br>5.333 | 0.033/0.002<br>16.5 |
| | 10m/s | 0.107/0.140<br>0.764 | 0.276/0.140<br>1.971 | 0.204/0.041<br>4.975 | 0.056/0.006<br>9.333 |
| broadleaf forest | 4m/s | 0.002/0.049<br>0.041 | 0.406/0.049<br>8.285 | 0.007/0.004<br>1.75 | 0.003/0.0001<br>30.0 |
| | 6m/s | 0.037/0.098<br>0.377 | 0.397/0.098<br>4.051 | 0.188/0.017<br>11.058 | 0.075/0.0008<br>93.75 |
| | 10m/s | 0.008/0.224<br>0.035 | 0.302/0.224<br>1.348 | 0.054/0.054<br>1.0 | 0.016/0.004<br>4.0 |

Minimum confidence scores of true detections/ Maximum confidence scores of false detections (average over all detections and all integral images). Values below 1 (red) indicate false classifications. The maximum·median combination performed best. In the case of occlusion (conifer and broadleaf forest), it separated false from true confidence scores by a factor of 4-93 (green).

Inter-process queues are applied for efficient and secure communication between sub-processes. The drone communication sub-process interacts with the drone, utilizing a MikroKopter-customized serial protocol to receive IMU/GPS positions and send waypoint instructions which include GPS location, orientation, and speed. Received GPS positions and orientations are time-stamped and communicated to the image-processing sub-process.

## A. Image Acquisition

For imaging, we grabbed digital video frames from the video capture card connected to the thermal camera using OpenCV's video capturing module. The camera was set to SAR mode (mode suited to search-and-rescue operations in the wilderness, using 100% field of view), providing a high tonal range for higher temperatures and fewer gray levels for colder temperatures. The images were time-stamped to assign individual GPS coordinates, preprocessed using OpenCV's pinhole camera model to remove the lens distortion, and cropped to a field of view of 43.10° and a resolution of 512 px × 512 px.

To match the slow and asynchronous GPS signal rate (5 Hz in our case) to the faster video rate (30 Hz in our case), we applied time-based linear interpolation (assuming constant flying speeds). We thus interpolated GPS coordinates for each video frame grabbed. This, however, led to an interpolation error that depends on flying speed and imaging time:

Given the drone's constant flying speed $v_f$[m/s] and the camera's imaging time $t_i$[s], the maximum interpolation error caused by an unknown delay between capturing and measuring the capturing time stamp (which includes the transmission time of the image from the camera to the processor) is:

$$E_{Imax} = (v_f \cdot t_i)/2, \tag{8}$$

and adds to the GPS error. Thus, for $t_i = 0.033$ s (30 fps), and $v_f = 1$ m/s, 4 m/s, 6 m/s, and 10 m/s, $E_{Imax} = 1.67$ cm, 6.67 cm, 10 cm, and 16.67 cm, for example. Note that assuming a constant drone speed results in $E_{Imax}$ being independent of the speed of the GPS sensor, as GPS positions of recorded camera frames can be linearly interpolated if they cannot be measured sufficiently fast. Only for non-constant speed segments (e.g., during acceleration and deceleration) are fast GPS samples beneficial for more precise piecewise linear or non-linear interpolations.

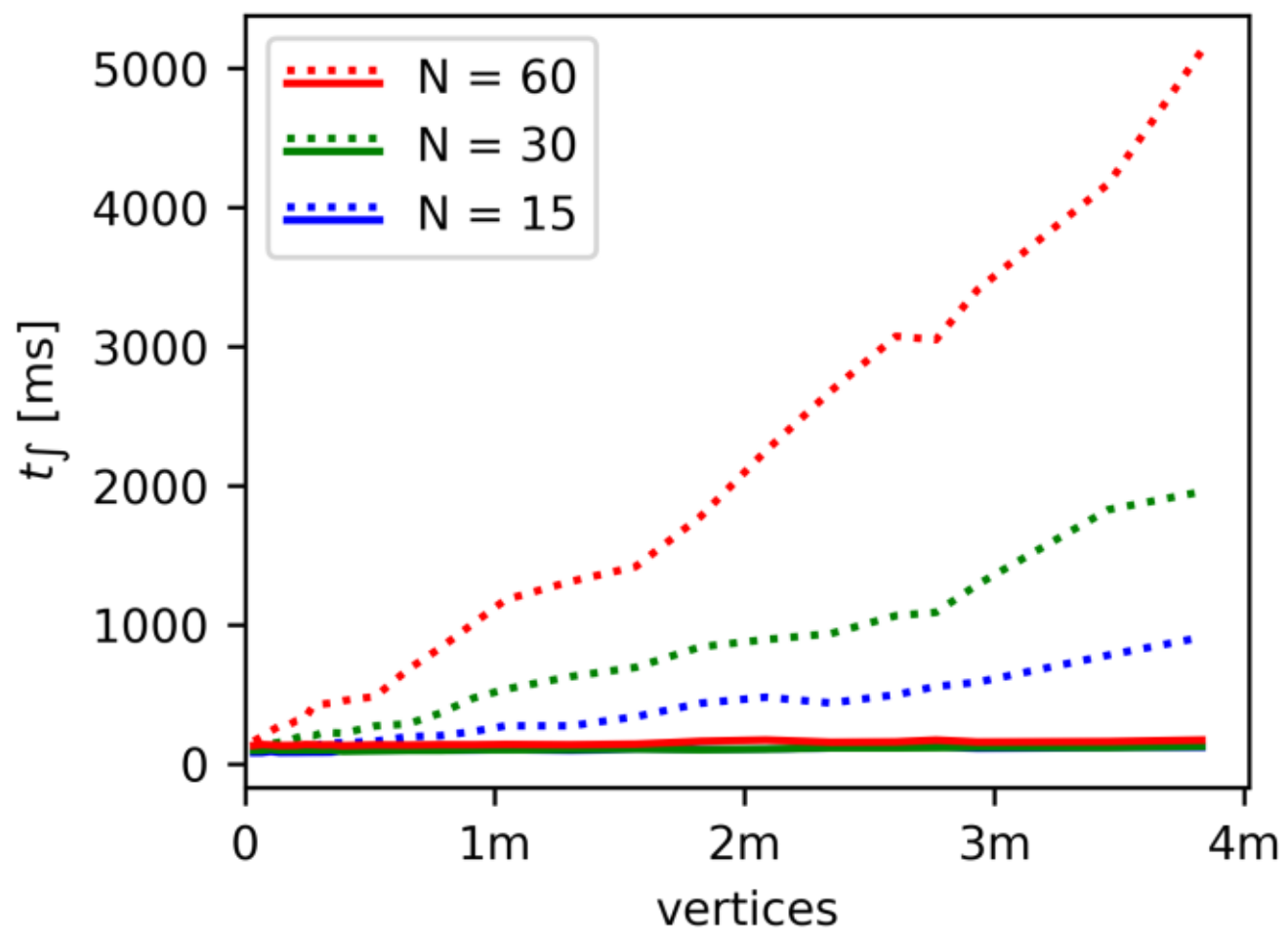

Fig. 7. Integration times [ms] for various N (number of images integrated) and a range of DEM sizes (vertex counts in millions). The new deferred rendering (solid lines) is compared with the old classical rendering (dashed lines). Classical and deferred renderings have similar computation times for small numbers of vertices. However, unlike for deferred rendering, rendering times drastically increase with vertex count for classical rendering. The plotted curves are averaged over 100 integral computations.

### B. Deferred Integral Imaging Computation

In principle, the preprocessed and GPS-assigned single thermal images are projected onto and averaged at a digital elevation model (DEM) using their individual poses and the camera's fixed intrinsic parameters. The DEM is a triangular mesh compatible with most standard graphics pipelines, as explained in [31]. For an integral image, the DEM with all projections is finally rendered from the center perspective of all integrated poses.

In [31], the above process was implemented with classical rendering, for which the entire DEM had to be processed for each projection. This did not achieve practical computation times for large numbers of projections and/or large DEMs (cf. Fig. 7). To decouple image projection from the processing of the DEM's geometry, we now utilize a rendering technique known as deferred rendering (DR). This requires the DEM's geometry to be processed only once for each integral image, and consequently speeds up computation time. With DR, the processed DEM geometry is preserved in the graphics memory and can be reused for all projections as long as the rendering perspective does not change. As shown in Fig. 7, compared to classical rendering, this leads to a significant decrease in computation time for large DEMs and large numbers of projections. Storing the DEM's processed geometry requires hardware support for floating-point precision buffers, which is supported by our SoCC.

Integral image computation was implemented using C, C++, and OpenGL, and integrated into the main Python program using Cython. Preprocessing and integration of 30 thermal images require around 99 ms and 115 ms, respectively, for DEMs with 34k-2.6m vertices.

### C. Classification

Person detection in the integral images was performed with the YOLOv4-tiny network architecture [56] on the VPU. More details on (pre-)training, parameters, and the selection of the best training weights can be found in [31].

Classifications are achieved in 84 ms and are optionally transmitted to a remote mobile device using the LTE communication hat connected to the SoCC.

For classifier training (using the Darknet software), we used 17 flights (F0-F11 and O1-O6; excluding F7) recorded in [30]. The data contains 7 flights over forests with persons on the ground (F0-F6), 4 flights over forests without persons (F8-F11), and 6 flights over a meadow with persons but without occlusions (O1-O6). From these flights F1, F8, and O6 were used for validation, while the remaining 14 flights were used for training. For our experiments, the recordings were resampled from 2D synthetic apertures to 1D synthetic aperture lines, pose data was computed from GPS readings, and persons were labeled, as explained in [31].

Note that manual compass correction was applied to each flight. To compute the integral images of the training and validation datasets, we applied the following augmentations: We varied the synthetic aperture size $N$ in 7 steps: 1 (pinhole), 5, 10, 15, 20, 25, and 30. Note that, due to resampling, some 1D apertures had fewer single images (e.g., the longest lines were only in the range $20 < N \leq 25$ for some scenes). Additionally, we applied 10 random image rotations by varying the up vector of the integral's virtual camera. The digital elevation model was translated up and down by 3 meters in steps of 1 meter to simulate defocus. Furthermore, we computed an additional integral image with a random compass error (rotation around each single-image camera's forward axis by +/- 15 degrees). This led to a total of 980 variations for each of the 179 resampled 1D apertures.

The trained network achieved an average precision score of 88.7% (without combined classification) on test data presented in Table 2 of [31] (previously 86.2% in [31]) and was also able to detect unoccluded persons in open terrain (e.g., meadows, fields), since it was additionally trained with the unoccluded person data from [30][57].

## VII. Conclusion and Future Work

We have shown that false detections can be significantly suppressed and true detections significantly boosted by classifying on the basis of multiple combined AOS - rather than single - integral images. This improves classification rates especially in the presence of occlusion. To make this possible, we modified the AOS imaging process to support large overlaps between subsequent integrals, thus enabling real-time and on-board scanning and processing of groundspeeds up to 10 m/s (while previously the maximum was 1 m/s [30][31]).

Due to the slow sampling rate of standard GPS, only constant-speed segments are currently supported. For acceleration and deceleration segments, faster and more precise Real Time Kinematic (RTK) devices are beneficial. Non-linear interpolation and prediction models will be investigated in the future to enable better mapping of faster imaging rates to slower pose sampling. Furthermore, we will investigate additional improvements in adaptive combination techniques and options for extending flight endurance to enable fully autonomous beyond-visual-line-of-sight (BVLOS) search-and-rescue missions.

## Appendix

Here we present the derivation of integrated occlusion density $\widetilde{D_\alpha}$ (7) for a ray passing through an occlusion volume at an angle $\alpha$ based on the statistical model described in [25]. Integrated occlusion density $\widetilde{D}$ for a ray of length $l$ passing orthogonally ($\alpha = 0°$) through the occlusion volume of density $D$, height $l$, and filled with occluders of uniform distribution and size $o$ is [25]:

$$\widetilde{D} = 1 - (1 - D)^{l/o}. \qquad (9)$$

The length of a ray passing at an oblique angle $\alpha$ ($\alpha > 0°$) through the same occlusion volume is $l/(cos(\alpha))$, and substituting this in (9) yields:

$$\widetilde{D_\alpha} = 1 - (1 - D)^{l/cos(\alpha)o}. \qquad (10)$$

Applying the logarithm after simplifying both (9) and (10) respectively leads to:

$$log\,(1 - \widetilde{D}) = \frac{l}{o} log\,(1 - D) \qquad (11)$$

and

$$log\,(1 - \widetilde{D_\alpha}) = \frac{l}{cos(\alpha)o} log\,(1 - D). \qquad (12)$$

Substituting (11) in (12) yields:

$$log\,(1 - \widetilde{D_\alpha}) = \frac{1}{cos(\alpha)} log\,(1 - \widetilde{D}). \qquad (13)$$

Transforming the logarithmic equation in (13) to its equivalent exponential equation and simplifying yields (7).

## Code Availability

All source code and data is available at: https://github.com/JKU-ICG/AOS.